# Explaining Aviation Safety Incidents Using Deep Temporal Multiple Instance Learning

Vijay Manikandan Janakiraman
USRA/ NASA Ames Research Center
Moffett Field, CA, USA
vjanakir@mail.nasa.gov

## ABSTRACT

Although aviation accidents are rare, safety incidents occur more frequently and require a careful analysis to detect and mitigate risks in a timely manner. Analyzing safety incidents using operational data and producing event-based explanations is invaluable to airline companies as well as to governing organizations such as the Federal Aviation Administration (FAA) in the United States. However, this task is challenging because of the complexity involved in mining multi-dimensional heterogeneous time series data, the lack of time-step-wise annotation of events in a flight, and the lack of scalable tools to perform analysis over a large number of events. In this work, we propose a precursor mining algorithm that identifies events in the multidimensional time series that are correlated with the safety incident. Precursors are valuable to systems health and safety monitoring and in explaining and forecasting safety incidents. Current methods suffer from poor scalability to high dimensional time series data and are inefficient in capturing temporal behavior. We propose an approach by combining multiple-instance learning (MIL) and deep recurrent neural networks (DRNN) to take advantage of MIL's ability to learn using weakly supervised data and DRNN's ability to model temporal behavior. We describe the algorithm, the data, the intuition behind taking a MIL approach, and a comparative analysis of the proposed algorithm with baseline models. We also discuss the application to a real-world aviation safety problem using data from a commercial airline company and discuss the model's abilities and shortcomings, with some final remarks about possible deployment directions.

## 1 INTRODUCTION

Explanations for Aviation safety incidents may be required for a multitude of reasons. A safety analyst working for a commercial airlines company may be interested in finding the root causes of incidents to improve the fleet's safety, to inform predictive maintenance, to monitor human factors such as fatigue, and situational awareness to improve pilot training etc. On the other hand, FAA and NASA often require such explanations to inform better airspace designs, improve policies, regulations and standard operating procedures. Organizations such as the National Transportation Safety Board (NTSB) may benefit from having explanations to safety incidents while conducting accident investigations.

Currently, safety incidents are explained manually; a group of human experts analyze a given safety incident and offer explanations using causal factors and correlated events that occurred in the flight. However, this is not a scalable approach with the numerous safety incidents that occur and with the growth in the volume of sensory data. In the US National Airspace alone, more than 9 million scheduled passenger flights operated in 2016 with an estimated 5000 overhead flights at any given time [2], generating large volumes of data at different levels in the airspace system. A commercial passenger airline with a fleet size of 280 operates about 1000 short to medium range flights per day, flags about 900 safety incidents[1]. While these numbers are expected to grow, it becomes impossible to manually analyze the numerous safety incidents on a case-by-case basis to come up with actionable insights to improve safety in a timely manner. This motivates us to develop a system that can analyze the flight data and offer explanations in an automated way.

We propose using precursors as a means to identify key events in the flight time series data. A precursor is any correlated event that occurs prior to the safety incident. Precursors give insights into the root causes of the incident and provide actionable insights with early alerts and corrective actions. To provide useful explanations, precursor mining aims to answer the following questions - "When do degraded states begin to appear?" "What are the degraded states?" "Are there corrective actions?" "What is the likelihood that the safety incident will occur?". For example, our method identified precursors to a particular high speed exceedance (safety incident) and helped explain the incident as follows. During the landing phase of the flight (see Figure 1), about 25 miles away from the runway, most variables were normal indicating a "safe state." Correspondingly the probability of the safety incident was close to zero. At about 2500ft altitude and 13 miles away, the flight made its turn to align with the runway when the speed reference was set incorrectly (unusually high) which caused the engine speed and consequentially the flight's speed to be high. The increase in probability of the safety incident indicates this as a precursor. At about 8 miles out, the speed reference was corrected (reduced) which caused the airspeed to drop. This is indicated by a decreasing probability of failure indicating possible corrective actions. However, from about 1500 ft altitude, the likelihood increases because the flaps were delayed. Flaps are friction surfaces that help reduce airspeed. In this case, flaps are not set on time which caused the airspeed to remain high. This happened when the flight approached the 1000 ft altitude marker that the safety incident was flagged. Thus, a 250 dimensional time series data was summarized to provide explanations about key events that occurred prior to the safety incident.

---

[1]The incident statistics are based on exceedance counts obtained from a de-identified airline company. The exceedances considered here fall under severity 2 and 3 while severity 1 is often ignored because of its mildness. Also note not all exceedances have equal safety concern.

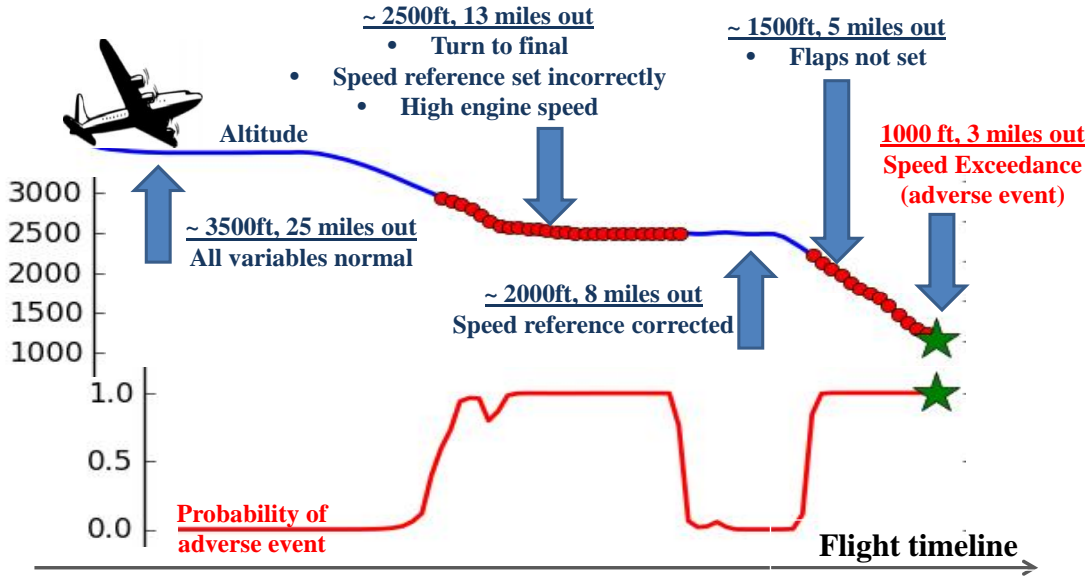

Figure 1: Explanations in the form of key events identified by precursor mining algorithm for a flight that had a high speed exceedance (safety incident). Important events are shown by blue arrows, precursors are marked with red circles while a green star is used to indicate the safety incident.

Precursor mining is a weakly-supervised learning problem. Usually we have easy access to a label that indicates the occurrence of a safety incident in a flight. The labels may come from aviation safety reporting system (ASRS) [28] where the flight crew members and air traffic controllers volunteer to report safety incidents during their flight, or from exceedance reports that are automatically generated by using domain based rules to identify safety incidents. While a flight-level label is available, the labels for precursors within a flight are usually not available, which makes the explanation task challenging. To address this, our past work focused on using the flight-level information as a weak supervision to obtain the expert's decision making behavior using inverse reinforcement learning [18]. In this paper, we propose to use Multiple-Instance Learning (MIL) to use the flight-level label to infer time-step wise event labels.

A MIL setting typically assumes a set of data instances grouped in the form of bags. The bag level labels are known while the instance labels are unknown. The task of MIL is to learn a model to either predict the bag labels or instance labels or both. An assumption is usually made in MIL that relates the instance-level labels to the bag-level labels. A standard assumption of MIL states that the bag is labeled positive if there is at-least one positive instance in the bag while it is labeled negative if all instances in the bag are negative. A more detailed discussion on MIL and the assumptions involved can be found in [5, 13, 14]. The standard MIL formulation does not consider time connection between instances and its performance drops when applied to time series data [15]. To address this, we propose a deep temporal multiple-instance learning (DT-MIL) model that extends the MIL framework to be applied to time series data. Further, by having deep neural network models, the approach can be scaled better for large data sets using GPU parallelism. To summarize, the main contributions of the paper include

(1) a novel deep temporal multiple-instance learning (DT-MIL) framework that combines multiple-instance learning with deep recurrent neural networks suitable for weakly-supervised learning problems involving time series or sequential data.
(2) a novel approach to explaining safety incidents using precursors mined from data.
(3) detailed evaluation of the DT-MIL model using real-world aviation data and comparison with baseline models.
(4) precursor analysis and explanation of a representative safety incident using multi-dimensional flight data from a commercial airline.

The rest of the paper is organized as follows. We discuss related work in Section 2. In Section 3, the precursor mining problem is formulated with an intuition about our approach followed by description of the DT-MIL model. The discussion about data, the considered safety incident and experiments are in Section 4. In Section 5, we present quantitative results of the DT-MIL method with a comparative analysis using relevant baselines and discuss the explanations offered for a few representative flights. This is followed by conclusions in Section 6.

## 2 LITERATURE REVIEW

The application of data mining and machine learning methods for aviation safety is not new. NASA Ames Research Center has been pioneering this line of work for several years and has developed methods such as Orca [8], i-Orca [9], Inductive Monitoring System [17], Multiple Kernel Anomaly Detection (MKAD) [11], ELM based anomaly detection [20], among others. For event detection and optimal alarm systems, NASA has open-sourced the Adverse Condition and Critical Event Prediction Toolbox (ACCEPT) which is a suite of machine learning algorithms [25]. Outside of NASA, much of the work revolves around finding anomalies in flight data [24, 29] or in text reports [4, 31]. Another related research is in understanding human factors in flight safety [23]. The above work aims at detecting unsafe patterns but offers little or no automated explanations.

Recently, several approaches for precursor mining have been proposed. The Automatic Discovery of Precursors in Time Series (ADOPT) finds precursors using multidimensional sensory time series data and has been applied to single flight incidents such as a take-off stall risk [19], as well as to incidents such as go-around [18] that involve interactions between multiple flights. Another approach is using a nested multiple-instance learning that finds precursors to protests and social events using text data [26]. Other story telling based algorithms exist, for instance [16], that are aimed at entity networks and using text documents but it is not trivial to see its applicability for multidimensional sensory data. Compared to the above methods, our approach combines multiple-instance learning (MIL) and deep recurrent neural networks (DRNN) to take advantage of MIL's ability to model weakly-supervised data and DRNN's ability to model long term memory processes, to scale well to high dimensional data and to large volumes of data using GPU parallelism.

Multiple-instance learning was first proposed by Dietterich et al. [13] which was then extended using support vector machines [5] and neural networks [27]. There exists some work on using multiple instance learning for temporal data such as using autoregressive hidden Markov models [15], and using recurrent neural networks [12] mainly to learn a prototype vector using the instances which is then used in the recurrent network for supervised classification of the bags. The motivation of this paper appears to be different from ours and it does not take advantage of the superior capabilities of long-term memory units such as the LSTM units. Only recently, deep multiple-instance learning has been proposed and used in the context of object detection and annotations [7, 30, 32]. Our work is an extension to the prior literature by combining multiple-instance learning with deep recurrent neural networks which appears to be novel.

## 3 METHODOLOGY

In this section, we add more formalism to the precursor mining problem extending our earlier work [18], discuss the idea behind using multiple-instance learning and introduce the deep temporal multiple-instance learning (DT-MIL) framework.

### 3.1 Precursor Mining Problem

The problem of precursor mining can be stated as follows. Let the time series data be represented by $X$ with dimensions $(N, L, d)$, where $N$ denotes the number of time series records, $L$ the maximum length of time series[2] and $d$ the number of sensory variables. Let

[2]variable length sequences are accommodated.

an instance in record $i$ at time $t$ be given by

$$e_t^i = f(\mathbf{x}_1^i, \mathbf{x}_2^i, ..., \mathbf{x}_t^i), \tag{1}$$

where $f(.)$ is some function that captures a transformed representation of the sequence $\mathbf{x}_1^i, \mathbf{x}_2^i, ..., \mathbf{x}_t^i$. Given the data $X_i$ and the corresponding safety incident label $Y_i$, the goal of precursor mining is to find a set of time instances $\mathcal{P}_i = \{j\}, 1 \leq j \leq L$ for which

$$P(\text{safety incident}|e_j^i) > \delta. \tag{2}$$

Note that the labels are binary; i.e., $Y^i = 1$ implies a safety incident reported in $X^i$. The label is also for the entire flight $X^i$ which is a collection of instances $e_1^i, e_2^i, ..., e_t^i, ..., e_L^i$. Each $\mathbf{x}_t^i$ is a vector of $d$ sensory variables that may include airspeed, altitude, engine states, pilot switch positions, autopilot modes among others.

### 3.2 Intuition

Following the definition from above, precursor mining is a weakly-supervised learning task. The only supervision comes from the summary label for a flight that says if it had a safety incident or not. Using this high-level information, the goal is to identify low-level events that occurred during the flight that correlate to the safety incident. Multiple-instance learning is a natural fit for weakly supervised problems where the bag labels (high-level supervision) are used to infer the instance labels (low-level). Further, a flight spans a finite time and events that happen during a flight are correlated temporally. For example, a flight typically involves dynamics at multiple time scales. An increase in throttle causes a response in engine speed in the order of milliseconds while it takes a few seconds to cause a notable increase in the airspeed of the flight. Also, when there is a change in wind condition or traffic pattern, it takes much longer to see its effect reflected in the flight's behavior. Thus it is important to consider instances in the flight along with a context history (seen in the definition of an instance $e_t^i$ in equation (1)). Figure 2 shows our proposed idea in which each flight is a multidimensional time series with a bag-level label that indicates if the flight had a safety incident or not. Here the instances are defined as the sequence of measurements up to the current time. In this way, an instance at time $t$ can be thought of as the event that occurs at time $t$, given the context up to that time. The standard MIL formulation does not capture temporal connection between data instances [15], and we propose a MIL framework that address this problem in the next section.

### 3.3 The DT-MIL Model

In this section, we develop the deep temporal multiple-instance learning model for precursor mining. Earlier, we defined an instance $e_t^i$ by using a history of data measurements. To efficiently model the instances, we use a specific type of recurrent neural networks involving the gated recurrent unit (GRU) [10]. Motivated by long short-term memory (LSTM) units, GRUs are proposed as a variant to simplify computation and implementation. Compared to LSTM which has a memory cell and four gating units that adaptively control the information flow inside the unit, the GRU has only two gating units [10]. Owing to its simplicity over LSTM units, we choose GRU as the recurrent units for our model. The information flow in a GRU is shown in Figure 3 while its governing equations

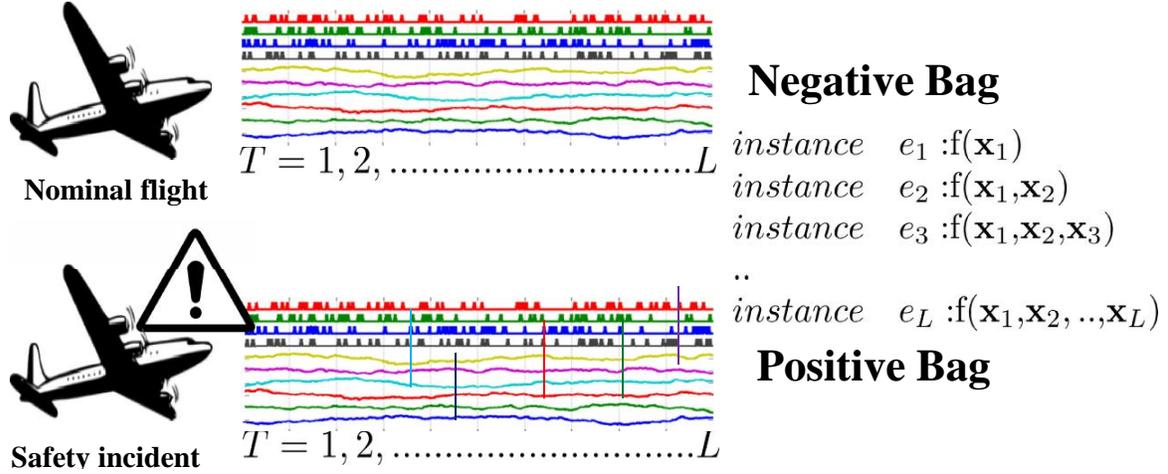

Figure 2: Figure showing the idea of a bag representation of a flight with instances corresponding to sequence data up to the current time. A flight is labeled positive if a safety incident is reported.

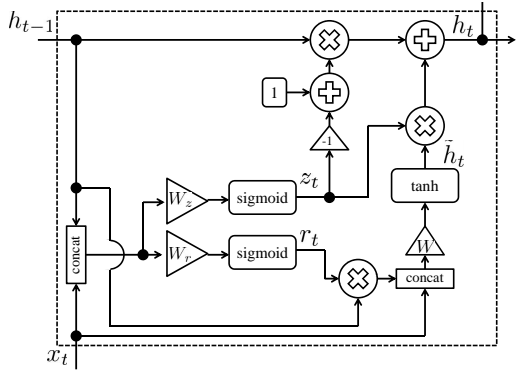

Figure 3: Figure showing the data flow in a gated recurrent unit (GRU).

are shown below

$$z_t = sigmoid(W_z.[h_{t-1}, x_t]) \quad (3)$$
$$r_t = sigmoid(W_r.[h_{t-1}, x_t]) \quad (4)$$
$$\tilde{h}_t = tanh(W.[r_t * h_{t-1}, x_t]) \quad (5)$$
$$h_t = (1 - z_t) * h_{t-1} + z_t * \tilde{h}_t. \quad (6)$$

The parameters of GRU include $W_r$, $W_z$ and $W$. The reset signal $r_t$ determines if the previous hidden state should be ignored while the update signal $z_t$ determines if the hidden state $h_t$ should be updated with the new hidden state $\tilde{h}_t$. By having many units each having its own reset and update signals, the GRU learns to capture the dependencies from past data over different time scales [10].

The DT-MIL architecture is shown in Figure 4. The time series data is processed by GRU units which convert the sequence of data into hidden states that are passed on to a layer of fully connected *tanh* units. The recurrent layer captures the temporal dependencies in the data while the fully connected layer adds more approximation capability to the model. Additional recurrent and fully connected layers may be added depending on the data complexity. This combination of recurrent and fully connected layers model the deep representation function $f(.)$ in the instance definition in equation (1). The instances $e_t$ are obtained after the fully connected layers which are then fed into a logistic layer to convert the instances into probabilities $p_1, p_2, .., p_t, .., p_L$. The MIL aggregation function $a(.)$ converts the instance probabilities into a bag probability $\hat{y}$ as

$$\hat{y}^i = a(p_1^i, p_2^i, .., p_t^i, .., p_L^i). \quad (7)$$

The loss function for learning the model parameters is a binary cross-entropy function given by

$$L(y, \hat{y}) = -[ylog(\hat{y}) + (1-y)log(1-\hat{y})]. \quad (8)$$

where $y$ represents the bag label. The above setup can be easily extended to a multi-class case by having many independent logistic layers followed by MIL aggregation layers for each class. It must be noted that multiple logistic layers may be better suited compared to a soft-max layer because a flight may have multiple safety incidents that may not be independent of each other. The model can then be trained using standard backpropagation taking advantage of all the recent advancements made in deep learning including GPU parallelism.

The MIL aggregation is an important function that mimics the MIL assumptions under the cross-entropy loss function. For example, if $a(.)$ is set to a $max(.)$, then it mimics the standard MIL assumption. To see how this works, consider a negative bag with a label of 0. To minimize the loss function, the $\hat{y}$ will be pushed close to 0. Given $\hat{y}^i = max(p_1^i, p_2^i, .., p_t^i, .., p_L^i)$ and when $\hat{y}$ is made small, the instance probabilities $p_1^i, p_2^i, .., p_t^i, .., p_L^i$ are all pushed close to 0 simultaneously. On the other hand, for a positive bag with a label of 1, with the aggregation function being a $max(.)$, at least one of the instance probabilities will be pushed close to 1 so that $\hat{y}$ becomes 1. Other aggregation functions such as mean, weighted mean, approximated max, Noisy-OR [33], Noisy-AND [22] may be used to reflect different MIL behavior and assumptions.

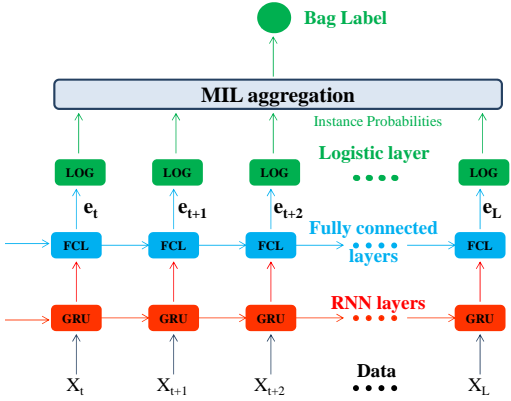

**Figure 4: Architecture of DT-MIL model showing the different layers.**

## 4 EXPERIMENTS

### 4.1 Data
For this study, we use the Flight Operational Quality Assurance (FOQA) data provided by a de-identified commercial airline[3] operated between April 2010 to October 2011 in Europe. The FOQA data consists of most of the sensory measurements on board the aircraft including flight speed, altitude, flight control surfaces, thrust, engine power, fuel consumption, pilot switches, pitch, roll, pressure, temperature among many others. The data is sampled at 1 Hz. Most of the data used in this work are recorded from flights that takeoff from these airports. To eliminate variability in aircraft characteristics, we consider only the Airbus models A319 and A320 because they belong to the same weight category of interest in this work.

### 4.2 Safety Incident
Accident statistics [3] show that about half of the fatal accidents occur during the last 4% of the flight - the final approach and landing. One of the key causes of landing related accidents involve energy mismanagement [1], which often lead to loss of control, tail-strike before reaching the runway, runway overrun, hard landing etc. The aircraft energy is a function of the weight, airspeed, altitude, thrust, lift and drag forces, which requires careful monitoring and control by the crew.

In this work, we consider a high-speed exceedance (HSE) during landing as the safety incident. The HSE is a rule-based definition for a safety incident that is widely used in operations to analyze safety. HSE is defined as

$$\text{airspeed}_{[altitude=1000ft]} > \text{target} + \text{tolerance} \quad (9)$$

where airspeed and target are recorded as part of the FOQA data. We modify the HSE rule to include some temporal context as follows

$$\text{airspeed}_{[context]} > \text{target}_{[context]} + \text{tolerance}; \quad (10)$$

where the temporal context includes one nautical mile of flight trajectory either side of the 1000 ft altitude checkpoint. By including a temporal context around the check-point helps in identifying the flights that truly had a sustained speed exceedance and ignores the ones that are momentarily caused by noisy data, winds etc. If equation (10) is satisfied, the flight is labeled positive as having the HSE safety incident. Note that the DT-MIL setup does not require the time-step location of the safety incident. As long as the safety incident is known to have occurred during the flight, it is labeled positive[4].

### 4.3 Model Setup
From the FOQA data, we selected a subset of sensory variables for this study, excluding the ones that were clearly not useful for explaining the HSE event. Some of the continuous variables that were chosen include airspeed, altitude, angle of attack, speed target, speed reference, aileron position, elevator position, rudder position, stabilizer position, engine speed, pitch angle, roll angle, accelerations along the 3 axes, vertical speed, aircraft gross weight. Some categorical variables include commands on flaps, speed brakes, landing gear, activations of autopilot, autothrottle and various modes of the autopilot, flight director and flight such as flight director engage status, location capture, altitude modes, thrust N1 mode, thrust EPR mode, vertical speed mode, TCAS and STALL status etc [19]. The FOQA data includes over 300 sensory variables out of which we chose a subset of about 60 variables based on both domain knowledge and automated feature selection using Granger causality [19].

In about 15 months of operational data, we had about 500 flights that had the HSE event. We considered the approach and landing phases of the flight starting approximately 25 nautical miles away from the runway. We sampled the flight sensory data at every quarter nautical mile of ground distance flown until touchdown. Thus, each flight record is about 100 time steps long. Note that the DT-MIL algorithm does not require the time series records to have the same lengths. We split the data randomly into training, validation and testing data in proportions 50%, 30% and 20% respectively of the total data. The data was normalized to have a zero mean and unit variance.

The DT-MIL model was prototyped using Tensorflow. The DT-MIL model architecture is shown in Fig 4. It includes a GRU recurrent layer with 20 units, a fully connected $tanh$ layer with 500 units, and a logistic layer to output the instance probabilities. A $max(.)$ MIL aggregation layer was used to mimic the standard MIL assumption. The model was trained using one GPU on a Dell Precision workstation with 64GB of RAM. The ADAM [21] optimizer with a learning rate of 0.002 was used. An $L_2$ regularization with coefficient 0.01 was used for all model parameters. The best model was identified by monitoring the performance on the validation data.

## 5 RESULTS AND DISCUSSION
### 5.1 Quantitative Results
This section evaluates the performance of the proposed DT-MIL model based on predictions at a bag (flight) level. As the speed

---

[3]Owing to proprietary nature of the work, we do not disclose the airline name and data. We are making efforts to make the code available to the public.

[4]For other safety incidents that may not be easily defined using simple thresholds such as lack of crew awareness and air-traffic controller miscommunication issues, the label may may be obtained from the Aviation Safety Reporting System reports or from other relevant sources.

exceedance is defined using airspeed, finding precursors in terms of airspeed offer no novel insights and thus it is removed from the analysis. We also eliminated variables that are highly correlated with airspeed such as the ground speed, stabilizer position, above stall margin etc to avid providing trivial information to the model to classify the two classes of flights. For a comparative evaluation, we include the following baseline models

(1) MI-SVM: This is a support vector machine model that is adapted for multiple-instance learning [6] following the standard MIL assumption that the bag is labeled positive if there is at-least one positive instance in the bag while it is labeled negative if all instances in the bag are negative. This is closest to the DT-MIL with a $max()$ aggregation function.
(2) MI-SVM (temporal): The above MI-SVM does not consider the temporal connection between instances in a bag, which is necessary for the flight time series data (refer Section 3.1). To enable the MI-SVM model to capture temporal behavior, we append each data instance to its past $N_H$ instances. The $N_H$ is a hyper-parameter which is set based on cross-validation.
(3) DT-MIL (no temporal): This model is obtained by removing the recurrent GRU units from the DT-MIL model. By doing so, we remove the model's ability to capture temporal behavior. This is used as a baseline to quantify the benefits of the GRU layer.
(4) DT-MIL (shallow): This model is obtained by removing the fully connected layers from the DT-MIL model. By doing so, we restrict the model's learning ability and evaluate the case where the model only has a recurrent layer without any benefit from the dense layers.
(5) ADOPT: The automatic discovery of precursors in time series (ADOPT) algorithm from our previous work [18, 19]. ADOPT works based on using reinforcement learning techniques to model precursors. Although ADOPT does not fall under the MIL setting and is not optimized for the bag-level accuracy, it is still possible to obtain the bag-level AUC from the instance level probability scores following the standard MIL assumption.

For each model, the hyper-parameters are optimized using cross-validation. The area under the receiver operating characteristic (AUC) is used as the evaluation metric.

It can be seen from the results in Table 1, that the proposed DT-MIL model was able to achieve a high accuracy in predicting the adverse event. A breakdown of the benefits from the recurrent and the dense layers can also be observed using the DT-MIL (no temporal) and DT-MIL (shallow) baseline models. It is not surprising to see that the data is temporal and the adverse event is defined over a time window which makes it difficult for the non-temporal models (DT-MIL (no temporal) and MI-SVM) whose accuracies are the lowest compared to their temporal counterparts. It is also interesting to note that DT-MIL (shallow) achieves similar accuracy as that of the full DT-MIL model. One reason could be that the considered adverse event is simple enough (function of two variables - airspeed and speed target) to be captured by a shallow model. Note that for a complex definition of an adverse event (defined using images or text), the deep structure would make a larger impact. The SVM models perform poorly on this data in spite of performing cross-validation based hyper-parameter tuning. A possible reason could be that the SVM models follow a heuristic approach to multiple instance learning [6] and could have ended up in a local optima. Although time-embedding is done for MI-SVM (temporal), it doesn't capture the temporal behavior of the flight data sufficiently well as compared to the DT-MIL model. The sophisticated GRU units in DT-MIL model appears to capture the temporal behavior in the data more effectively. ADOPT performs poorly in predicting the bags which is expected. While it is not fair to compare ADOPT with the other MIL algorithms, we still wanted to have it as a baseline because ADOPT was our previous work on precursor discovery and is relevant to the problem considered in this paper.

Table 1: Area under the ROC curve (AUC) for bag-level predictions by DT-MIL compared against baseline models.

| Models | Train AUC | Test AUC |
|---|---|---|
| **DT-MIL** | **0.9904** | **0.9837** |
| DT-MIL (no temporal) | 0.9723 | 0.9447 |
| DT-MIL (shallow) | 0.9892 | 0.9789 |
| MI-SVM | 0.8052 | 0.79825 |
| MI-SVM (temporal) | 0.8751 | 0.8802 |
| ADOPT | 0.9373 | 0.8280 |

## 5.2 Instance Probabilities

From the trained DT-MIL model, we can obtain the instance level probabilities by recording the inputs to the MIL aggregation layer (see Figure 4). Using the instance probabilities, the significant events that explain the HSE incident may be identified. Figure 5 shows the instance probabilities for nominal flights and flights with the HSE event for the unseen test data. It can be seen that most of the flights (in both sub-figures) start the approach and landing phases from a relatively "safe" state with adverse event probabilities close to zero, which is expected. Instances that are about 25 nautical miles from touchdown may be too far out to show significant precursors to the HSE incident. As the flights approach the 1000 ft altitude instant where the HSE event is defined, the instance probabilities increase close to 1 (100% likelihood of HSE occurring) for the flights with HSE event (in red), indicating that the instances have information correlating to the HSE event. Such events may be considered as precursors in the explanation task. On the other hand, for the nominal flights (in green), the instance probabilities remain close to zero most of the times, indicating absence of precursors in majority. With the choice of a $max(.)$ function for the MIL aggregation, if the instance probability goes above 0.5 at any time during the flight, it will be considered a positive flight. This happens for a few nominal flights which accounts for the false positives. The instances that have a high probability indicate possible precursors even in nominal flights. However, for nominal flights, such precursors are usually followed by corrective actions which decrease the probability close to zero before the HSE checkpoint at 1000ft altitude (see discussion that relate to Figure 9). In other cases, there are false positives too (see discussion that relate to Figure 10). In the next section, we explain the HSE events using the precursors and corrective actions inferred from the instance probabilities.

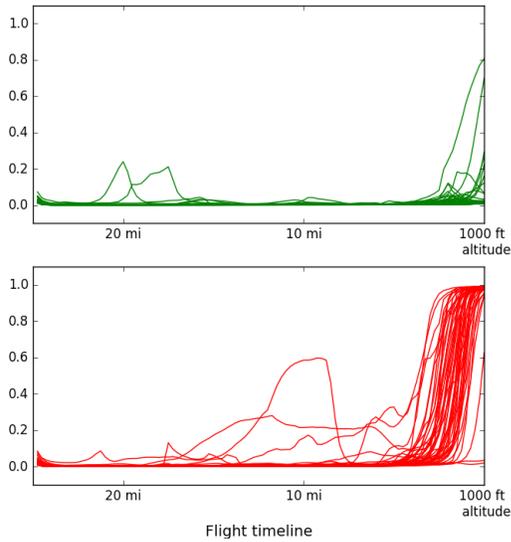

**Figure 5: Instance probabilities of nominal flights (in green) and flights with HSE safety incident (in red). Each subfigure shows about 50 flights. The probabilities are plotted against the flight timeline from 25 nautical miles from the runway (left end of the plot) to the HSE event at 1000 ft altitude checkpoint (right end of the plot).**

### 5.3 Safety Incident Explanation

Using DT-MIL, we analyze two flights that had the HSE incident and two nominal flights that were wrongly classified as a HSE flight (false positives).

Figure 7 shows a flight that had the HSE incident. In this flight, up to an altitude of 2000 ft (about 10 miles away from the runway), most variables are within nominal bounds indicating a "safe" state. When the flight reached about 1500ft altitude and about 5 miles away from the runway, the usual trend is to set the flaps. The flaps are friction surfaces that reduce the airspeed. However, this was delayed which caused the airspeed to remain higher causing the exceedance. In fact, based on the precursor insights, a further analysis confirmed that the flap setting variable is highly indicative of the high speed exceedance. Figure 6 shows the distributions of the time instants when final-flaps are set for nominal flights (colored green) against the flights that had the HSE event (colored in red). About 85% of the nominal flights have the final flaps set one nautical mile before the 1000 ft altitude checkpoint while only 12% of the HSE flights have the final flaps set at this time. Although it is not clear if flap setting causally influence the airspeed but the strong correlation emerged from the precursor analysis.

Figure 8 shows another HSE flight that was explained using DT-MIL precursors. Until about 15 miles away from the runway, the variables were nominal, again indicating a nominal or a "safe" state. Towards the end of the flight, it can be seen that the flaps were delayed similar to the previous example. This was detected as precursor by DT-MIL. However, this flight is different from the previous one in that there are earlier precursors that may have caused the airspeed to remain high even before the time of flap setting. Indeed, a higher setting of speed reference and a corresponding high engine speed were also detected as precursors by the DT-MIL model when the flight was about 10 miles away and at an altitude close to 3500 ft. This caused the airspeed to remain high along with indications of the precursor score increasing. The precursor variables are plotted along with altitude and airspeed indicating that at this point, the speed reference was higher than nominal which caused the engine speed to be high. This was then corrected when the flight traveled about 3 miles closer to the runway, at which point the autopilot was engaged. In spite of the correction in engine speed, the airspeed continued to be high at the usual time of flap setting. Thus it is possible that flaps were delayed because of the high airspeed that was caused by the earlier precursors. Although airspeed was removed from model training, the time instances of high airspeed both at the safety incident as well as previous instances were correctly identified by the model. The precursor score acts as a summary indicator of the risk factors that evolve in the flight over time, indicates these time instances well.

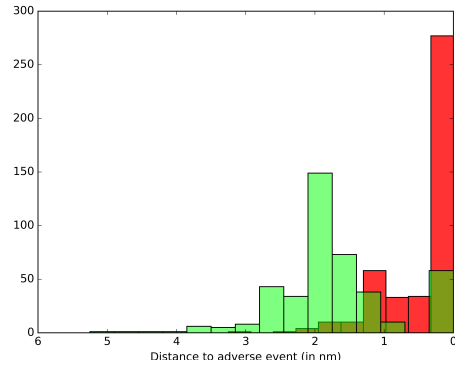

**Figure 6: Distribution showing early setting of final-flaps for nominal flights (colored green) against the flights that had the HSE event (colored in red).**

While false positive rates are low (see Table 1, we analyze the flights to understand model shortcomings. Figure 9 shows a nominal flight (no speed exceedance) where a precursor was seen and later corrected. At about 4000 ft altitude and 10 miles away from the runway, the flight had some mild precursors. It appears that the speed reference was set higher than nominal which caused the engine speed to remain high, in turn causing the airspeed to remain high momentarily. This was later corrected after the autopilot was engaged which caused these variables to reduce to nominal values. Although this flight is similar to the previous example in Figure 8, here there is significant correction with no further precursors such as late flaps which possibly prevented the safety incident. Although this flight was flagged as a false positive, the instance probabilities still offer some value in explaining the HSE related events that were observed in the flight.

Finally, we analyze another false positive flight that is truly a model error. Figure 10 shows a nominal flight whose variables were nominal until about 2000ft altitude and 4 miles away from the runway. However, close to the checkpoint, the final flaps were not set. Although this flight is similar to the first example (Figure 7), there is no speed exceedance. In this flight, although the flaps were delayed, the airspeed remains nominal in spite of the final flaps not set. The DT-MIL model finds this as a high-risk precursor and gives a high probability value although the HSE event is not flagged in this flight.

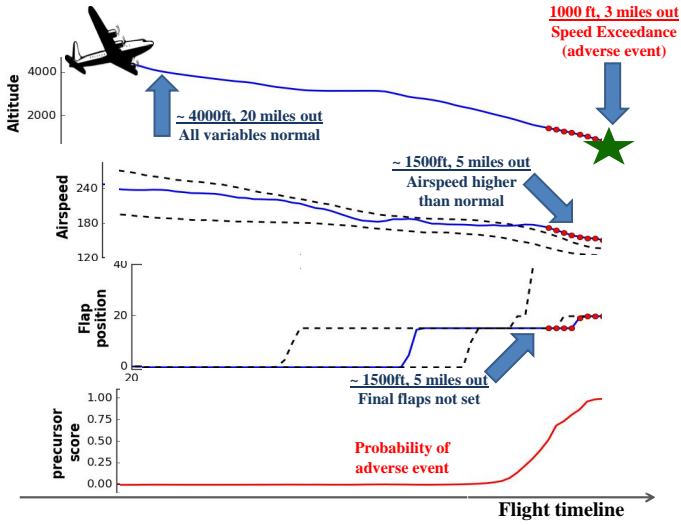

Figure 7: Explanation for a flight that had a HSE incident indicating a late flap setting precursor. Important events are shown by blue arrows, precursors are marked in red while a green star is used to indicate the HSE event.

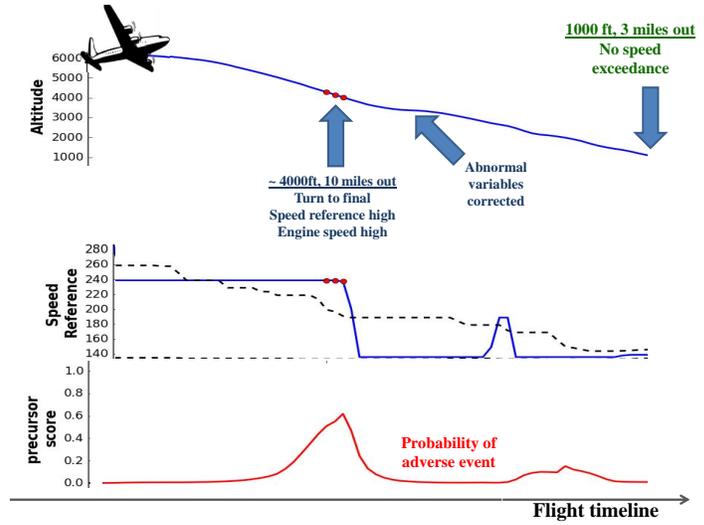

Figure 9: Explanation for a false positive flight involving a precursor followed by a correction.

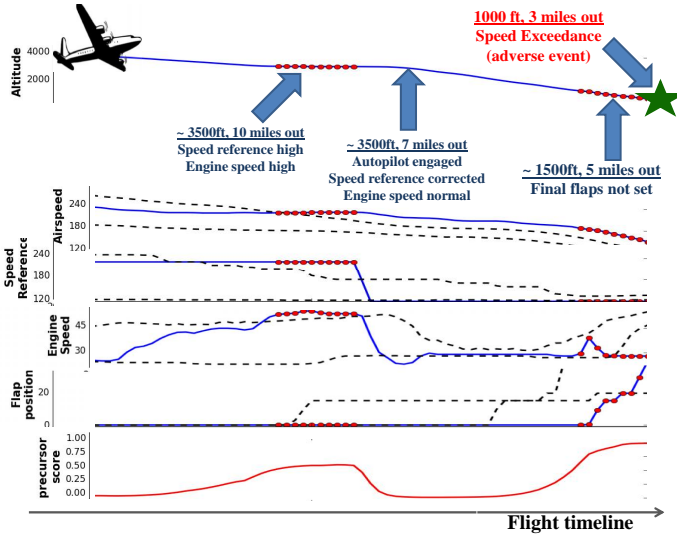

Figure 8: Explanation for a flight that had a HSE incident indicating a late flap setting precursor as well as an early high engine speed precursor. Important events are shown by blue arrows, precursors are marked in red while a green star is used to indicate the HSE event.

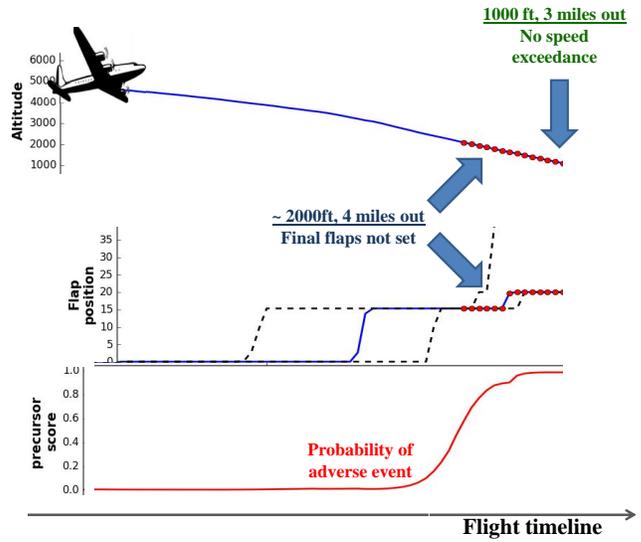

Figure 10: Explanation for a false positive flight where the model wrongly identified late flaps as precursors.

## 6 CONCLUSION

This paper introduces a precursor mining framework for explaining aviation safety incidents using flight recorded data. The DT-MIL algorithm extends multiple instance learning to temporal data using deep recurrent neural nets as building blocks. While the model is general enough to be applied to find precursors to any event of interest (even outside of aviation safety), this paper demonstrates the application using a high speed exceedance safety incident observed in a real commercial airline data.

While this work provides early insights into finding precursors and event explanations, further work must be done to fine-tune the models to make it suitable for deployment. Some of the possible directions for deployment include (1) developing a decision support system for the flight crew, air-traffic controllers and others in the real-time decision making loop for flight operations and (2) developing a data analytics pipeline to allow safety analysts and airline personnel to make use of the DT-MIL model to conduct a post-hoc analysis of flights to feed into pilot training or for designing operating procedures. For either of the tasks, our future work will involve extending the DT-MIL framework to analyze multiple safety incidents and formulate system-level safety margins to evaluate risks and risk mitigating maneuvers. Further, the models need to be re-calibrated to find operationally significant precursors by taking expert feedback.

## 7 ACKNOWLEDGMENTS

This research is supported by the NASA Airspace Operation and Safety Program. The authors would like to thank the subject matter experts Michael Stewart, Bryan Matthews and Robert Lawrence for their insightful comments and perspectives on the identified precursors.